%% file: main.tex
\documentclass[11pt,a4paper]{article}
\usepackage[hyperref]{acl2021}
\usepackage{times}
\usepackage{latexsym}

\usepackage{microtype}

\aclfinalcopy 
\def\aclpaperid{3196} 

\input{math_commands.tex}

\usepackage{microtype}
\usepackage{graphicx} 
\usepackage{float} 
\usepackage{hyperref}
\usepackage{url}
\usepackage{booktabs}
\usepackage{multirow}
\usepackage{graphics}
\usepackage{amsmath}
\usepackage{multicol}
\usepackage{lipsum}
\usepackage{mwe}
\usepackage{hyperref}
\usepackage{url}
\usepackage[caption=false]{subfig}
\usepackage{xcolor}
\usepackage[ruled,vlined]{algorithm2e}
\usepackage{bbm}
\usepackage{upgreek}

\SetAlFnt{\small}
\usepackage{wrapfig}
\newcommand{\sectionref}[1]{\S\ref{#1}}

\definecolor{asparagus}{rgb}{0.53, 0.66, 0.42}

\title{Learning Contextualized Knowledge Structures \\ for Commonsense Reasoning}

\author{Jun Yan$^{1}$, Mrigank Raman$^{2}$, Aaron Chan$^{1}$, Tianyu Zhang$^{3}$, Ryan Rossi$^{4}$, \\ \textbf{Handong Zhao$^{4}$, Sungchul Kim$^{4}$, Nedim Lipka$^{4}$, Xiang Ren$^{1}$} \\
University of Southern California$^{1}$, IIT Delhi$^{2}$, Tsinghua University$^{3}$, \\Adobe Research$^{4}$\\
\small{\texttt{\{yanjun, chanaaro, xiangren\}@usc.edu}}, ~\small{\texttt{mt1170736@iitd.ac.in}}, \\ \small{\texttt{zhang-ty17@mails.tsinghua.edu.cn}}, ~\small{\texttt{\{ryrossi, hazhao, sukim, lipka\}@adobe.com}}
}

\begin{document}

\maketitle
\begin{abstract}
\input{sections/0_abstract}
\end{abstract}
\input{sections/1_introduction}
\input{sections/2_problem}

\input{sections/3_method}
\input{sections/4_experiments}
\input{sections/5_related}

\input{sections/6_conclusion}
\input{sections/7_acknowledgements}

\bibliography{acl2021_rebiber}
\bibliographystyle{acl_natbib}

\clearpage
\appendix
\input{sections/8_appendix}

\end{document}

%% file: math_commands.tex
\usepackage{amsmath,amsfonts,bm}

\def\eqref#1{equation~\ref{#1}}

\def\1{\bm{1}}

\DeclareMathAlphabet{\mathsfit}{\encodingdefault}{\sfdefault}{m}{sl}
\SetMathAlphabet{\mathsfit}{bold}{\encodingdefault}{\sfdefault}{bx}{n}



%% file: sections/0_abstract.tex
Recently, knowledge graph (KG) augmented models have achieved noteworthy success on various commonsense reasoning tasks.
However, KG edge (fact) sparsity and noisy edge extraction/generation often hinder models from obtaining useful knowledge to reason over.
To address these issues, we propose a new KG-augmented model: Hybrid Graph Network (HGN).
Unlike prior methods, HGN learns to jointly contextualize extracted and generated knowledge by reasoning over both within a unified graph structure.
Given the task input context and an extracted KG subgraph, HGN is trained to generate embeddings for the subgraph's missing edges to form a ``hybrid" graph, then reason over the hybrid graph while filtering out context-irrelevant edges.
We demonstrate HGN's effectiveness through considerable performance gains across four commonsense reasoning benchmarks, plus a user study on edge validness and helpfulness.\footnote{Our code and data can be found at \url{https://github.com/INK-USC/HGN}.}


%% file: sections/1_introduction.tex
\section{Introduction} \label{sec:intro}

\begin{figure}[t]
\includegraphics[scale=0.43]{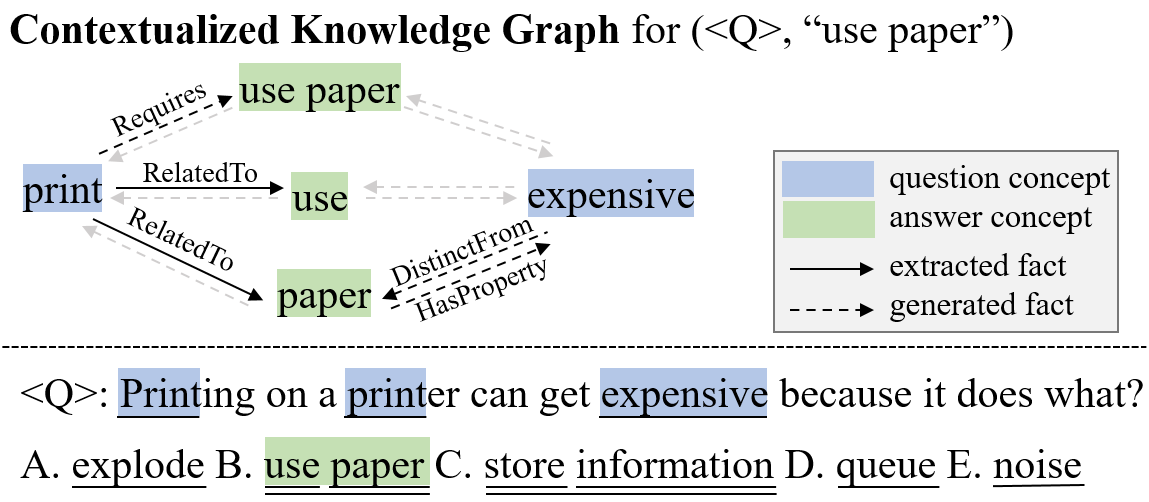}
\caption{\textbf{KG-Augmented Commonsense QA.} Predicting the correct answer (``use paper") requires commonsense facts like \begin{small}\texttt{(print, Requires, paper)}\end{small} and \begin{small}\texttt{(paper, HasProperty, expensive)}\end{small}, which are not given in the question and candidate answers. HGN uses facts extracted from the KG, \textit{e.g.}, \begin{small}\texttt{(print, RelatedTo, use)}\end{small}, but also generates new facts, eventually upweighting relevant ones, \textit{e.g.}, \begin{small}\texttt{(print, Requires, use paper)}\end{small} and \begin{small}\texttt{(paper, HasProperty, expensive)}\end{small}, while downweighting irrelevant ones, e.g., \begin{small}\texttt{(use, ?, expensive)}\end{small}.} 
\label{fig:example}
\end{figure}

Commonsense reasoning (CSR) is essential for natural language understanding (NLU) systems to function effectively in the real world \cite{apperly2010mindreaders}. 
For example, to answer the question in Figure~\ref{fig:example}, one must already know that \textit{printing requires using paper}. 
Yet, since commonsense knowledge is self-evident to humans, it is rarely stated in natural language \cite{gunning2018machine}. 
This makes it hard for neural pre-trained language models (PLMs) \citep{devlin2018bert} to learn commonsense knowledge from corpora alone \cite{marcus2018deep}.


Unlike raw text corpora, knowledge graphs (KGs) can provide structured commonsense facts (edges) of the form \begin{small}\texttt{(concept1, relation, concept2)}\end{small} \cite{speer2017conceptnet}. 
Hence, many recent CSR models augment the PLM with a KG, allowing such \textit{KG-augmented models} to make predictions via multi-hop reasoning over the KG \cite{lin2019kagnet, bosselut2019dynamic}.

Despite the growing success of KG-augmented models, obtaining helpful KG facts for a given task instance remains challenging. 
Existing models assume using either KG-extracted edges \cite{lin2019kagnet, ma2019towards, feng2020scalable, yasunaga2021qa}, PLM-generated edges (to address KG edge sparsity) \cite{bosselut2019dynamic}, or a late fusion of both \cite{wang2020connecting} is sufficient. Both extraction and generation can produce unhelpful edges, so the model must decide which edges to focus on during reasoning. 
Since extracted and generated edges are derived from the same set of concepts (nodes), modeling the interactions between extracted and generated edges jointly within a shared KG structure could provide stronger signal for identifying contextually relevant edges. However, current models do not leverage this information.


In response, we propose a new KG-augmented model: \textbf{Hybrid Graph Network (HGN)}.
Unlike prior models, HGN learns to jointly contextualize extracted and generated knowledge by reasoning over \textit{both} within a unified graph structure.
Given the task input (i.e., context) and an extracted KG subgraph, HGN is trained to generate embeddings for the subgraph's missing edges to form a ``hybrid" graph, then reason over the graph (to update model parameters) while filtering out context-irrelevant edges.
HGN achieves this primarily through edge reweighting, which downweights irrelevant edges, and edge-weighted message passing, which attenuates irrelevant edges' impact on reasoning.



Our extensive experiments demonstrate that HGN improves performance over all baselines across four CSR benchmarks. In particular, among comparable methods, HGN ranks first on the CommonsenseQA \citep{talmor2019commonsenseqa} and OpenbookQA \citep{mihaylov2018can} leaderboards. Plus, our user studies show that humans find HGN-filtered edges to be more valid and helpful than the heuristically extracted edges used in prior work.

%% file: sections/2_problem.tex
\section{Problem Statement}
\label{sec:problem}

We consider CSR tasks, like question answering (QA), which can benefit from commonsense KGs. To solve CSR tasks, we focus on KG-augmented models, where a PLM is augmented with a commonsense KG. Given a CSR task, let $x$ be the task's text input, $f$ be the model, and $f(x)$ be the model output. We denote a KG as $\mathcal{G}=(\mathcal{V},\mathcal{R},\mathcal{E})$. $\mathcal{V}$, $\mathcal{R}$, and $\mathcal{E}$ are the sets of nodes (concepts), relations, and edges (facts), respectively, in the KG. An edge is a directed triple of the form $e = (h,r,t)\in\mathcal{E}$, where $h\in\mathcal{V}$ is the head node, $t\in\mathcal{V}$ is the tail node, and $r\in\mathcal{R}$ is the relation between $h$ and $t$.
Let $[\cdot,\cdot]$ denote concatenation of text or vectors. 

\begin{figure}[t]
\vspace{-0.0cm}
\centering
\includegraphics[scale=0.4]{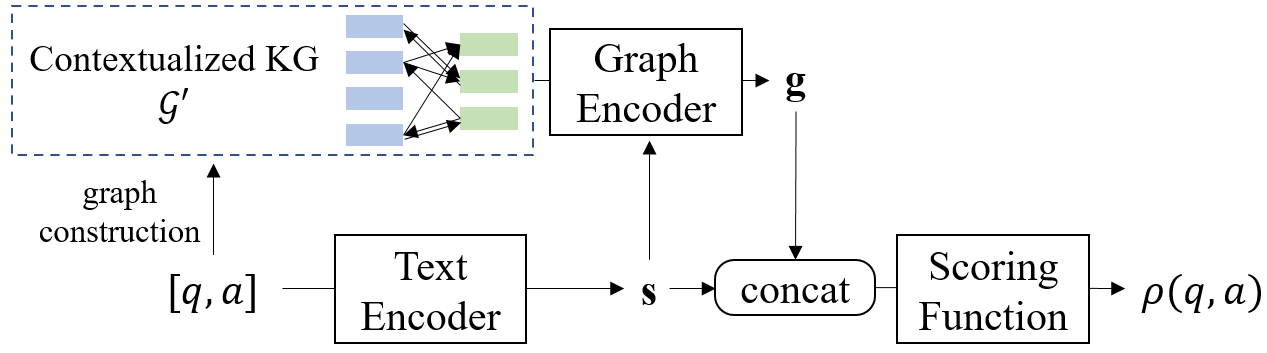}
\caption{\textbf{High-level schematic of a typical KG-augmented model for CSR.} In KG-augmented models, text encoder $f_{\text{text}}$ tends to be a Transformer PLM, and scoring function $f_{\text{score}}$ is usually an MLP. Meanwhile, KG-augmented models generally vary more in their graph encoder $f_{\text{graph}}$ and graph construction.}
\label{fig:typical}
\end{figure}

As illustrated in Figure~\ref{fig:typical}, a KG-augmented model $f$ has three main components: text encoder $f_\text{text}$, graph encoder $f_\text{graph}$, and scoring function $f_\text{score}$. First, $\mathbf{s}=f_\text{text}(x;\boldsymbol{\uptheta}_\text{text})$ is the encoding of $x$, where $f_\text{text}$ is usually a Transformer PLM. Second, as supporting evidence, a $x$-specific graph $\mathcal{G}'=(\mathcal{V}',\mathcal{R}',\mathcal{E}')$ is constructed from $\mathcal{G}$ (Figure~\ref{fig:example}). Typically, this is done via heuristic extraction by selecting $\mathcal{V}' \subseteq \mathcal{V}$ as the concepts mentioned in $x$, $\mathcal{R}' \subseteq \mathcal{R}$ as the relations between concepts in $\mathcal{V}'$, and $\mathcal{E}' \subseteq \mathcal{E}$ as the edges involving $\mathcal{V}'$ and $\mathcal{R}'$. If $\mathcal{G}$ does not provide enough knowledge to build a good $\mathcal{G}'$, then new edges are sometimes added to $\mathcal{G}'$ using a PLM-based generator \cite{wang2020connecting}. We call $\mathcal{G}'$ the contextualized KG. $\mathbf{g}=f_\text{graph}(\mathcal{G}', \mathbf{s};\boldsymbol{\uptheta}_\text{graph})$ is then the joint encoding of $\mathcal{G}'$ and $\mathbf{s}$. Third, the model output is computed as $f(x) = f_\text{score}([\mathbf{s}, \mathbf{g}];\boldsymbol{\uptheta}_\text{score})$, where $f_\text{score}$ is usually a multilayer perceptron (MLP). 
Existing KG-augmented models mainly differ in their design of $f_\text{graph}$, reasoning over the KG through message passing \cite{schlichtkrull2018modeling, feng2020scalable, yasunaga2021qa} or edge/path aggregation \cite{lin2019kagnet, bosselut2019dynamic, ma2019towards}.

While KG-augmented models can be applied to any CSR task involving KGs (e.g., natural language inference), we consider multi-choice QA in this work. Given a question $q$ and set of candidate answers $\{a_i\}$, the QA model's goal is to predict a plausibility score $\rho(q, a)$ for each $a \in \{a_i\}$, so that the highest score is predicted for the correct answer. To use KG-augmented models for commonsense QA, we set $x = [q, a]$ and $\rho(q, a) = f(x)$.

%% file: sections/3_method.tex
\section{Hybrid Graph Network (HGN)}
\label{sec:approach}





\begin{figure*}[t]
\centering
\includegraphics[scale=0.45]{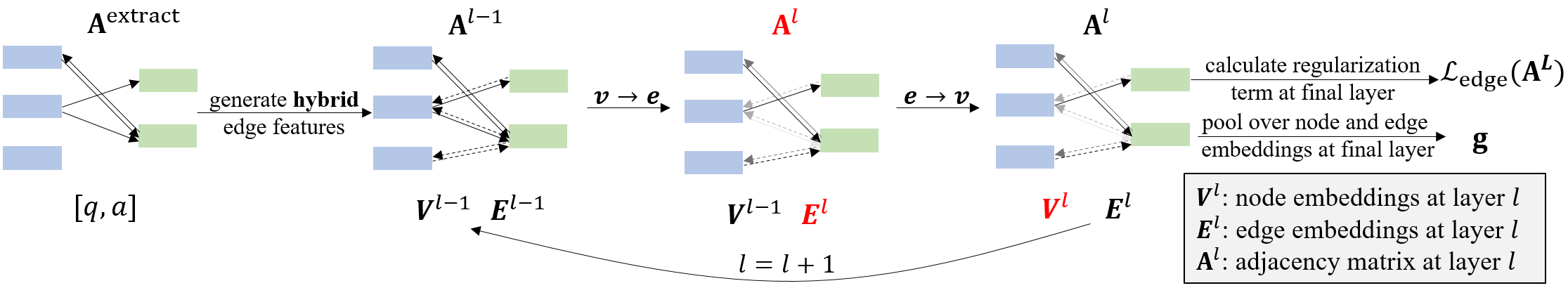}
\caption{\textbf{Overview of HGN.} After building a hybrid graph of extracted and generated edges (\sectionref{subsec:graph_construction}), HGN reasons over the hybrid graph by updating the node embeddings $\mathbf{V}$, hybrid edge embeddings $\mathbf{E}$, and adjacency matrix $\mathbf{A}$ at each layer $\ell$ (\sectionref{subsec:graph_reasoning}). \textbf{Darker} edges indicate higher weights. \textcolor{red}{\textbf{Red}} variables are updated in the previous step.}
\label{fig:hgn}
\end{figure*}

\subsection{Overview}
As illustrated in \sectionref{sec:problem} and Figure~\ref{fig:typical}, given question-answer pair $(q, a)$ for an instance of the multi-choice QA task, the KG-augmented QA model first obtains a $(q, a)$-contextualized KG $\mathcal{G}'$ via the full KG $\mathcal{G}$. Edges in $\mathcal{G}'$ can be extracted directly from $\mathcal{G}$ or generated using a PLM-based generator \cite{wang2020connecting, bosselut2019comet}. Then, the model transforms $(q, a)$ and $\mathcal{G}'$ into text encoding $\mathbf{s}$ and graph encoding $\mathbf{g}$, respectively. Finally, $\mathbf{s}$ and $\mathbf{g}$ are used to predict $(q, a)$'s plausibility.

However, a contextualized KG may have low knowledge recall or precision, hindering the QA model's access to relevant knowledge.
Low recall can stem from missing edges in $\mathcal{G}$, low precision can be the result of bad annotations in $\mathcal{G}$, and both can be caused by noisy edge extraction or generation when building $\mathcal{G}'$.
HGN addresses these issues by reasoning over both extracted and generated edges within a unified graph structure.
To improve recall, HGN generates new edges via a PLM-based generator, then initializes a hybrid contextualized KG containing both extracted and generated edges. Note that edge generation is generally $(q, a)$-agnostic and may produce irrelevant edges that hurt knowledge precision. 
To improve precision, HGN learns to reweight edges in the hybrid graph and reason over the hybrid graph via edge-weighted message passing. 
This is akin to learning the hybrid graph's structure and reduces the impact of irrelevant edges on reasoning. 
Additionally, to further encourage downweighting of noisy edges during reasoning, HGN is trained with entropy regularization on the learned edge weights.

The overall learning objective of HGN is defined as $\mathcal{L}=\mathcal{L}_\text{task} + \beta\mathcal{L}_\text{edge}$, where $\mathcal{L}_\text{task}$ is the loss for the downstream task (in our work, QA), $\mathcal{L}_\text{edge}$ is the entropy regularization term for edge weights, and $\beta \geq 0$ is a loss weight hyperparameter.
In the following subsections, we first explain how the contextualized KG $\mathcal{G}'$ is constructed as a hybrid graph, including its node embeddings $\mathbf{V}$, hybrid edge embeddings $\mathbf{E}$, and adjacency matrix $\mathbf{A}^0$ (\sectionref{subsec:graph_construction}).
Next, we show how HGN uses edge-weighted message passing to update $\mathbf{V}$, $\mathbf{E}$, and $\mathbf{A}^0$ for $L$ layers (Figure~\ref{fig:hgn}), yielding a refined adjacency matrix $\mathbf{A}^L$ of learned edge weights (\sectionref{subsec:graph_reasoning}). 
Finally, we describe how $\mathcal{L}_\text{task}$ is computed using $\mathbf{s}$ and $\mathbf{g}$, while $\mathcal{L}_\text{edge}$ is calculated using $\mathbf{A}^L$ (\sectionref{subsec:learning_objective}).

\subsection{Hybrid Graph Construction}
\label{subsec:graph_construction}




\paragraph{Node Embeddings.}
The first step of retrieving knowledge from $\mathcal{G}$ is concept grounding, which involves identifying text spans in $(q, a)$ that match nodes in $\mathcal{V}$. We define $\mathcal{V}'$ as the set of all concepts mentioned in $(q, a)$, where $\mathcal{V}'_q=\{v_i\}_{i=1}^{n_q}$ and $\mathcal{V}'_a=\{v_i\}_{i=1}^{n_a}$ are the question and answer concepts, respectively. Each node $v_i \in \mathcal{V}'$ is represented by an embedding $\mathbf{v}_i \in \mathbf{V}$, which can be initialized using BERT \cite{devlin2018bert} or TransE \cite{bordes2013translating}. 


\paragraph{Hybrid Edge Embeddings.}
In $\mathcal{G}'$, we loosen the definition of an edge to be $e_{(i,j)} = (v_i, v_j) \in \mathcal{E}'$.
We build fully-connected edges between question and answer nodes in $\mathcal{G}'$.
The set of edges in $\mathcal{G}'$ is thus defined as $\mathcal{E}' = (\mathcal{V}'_{q}\times \mathcal{V}'_{a}) \cup (\mathcal{V}'_{a}\times \mathcal{V}'_{q})$. After concept grounding, we need an edge embedding $\mathbf{e}_{(i,j)} \in \mathbf{E}$ for each edge $e_{(i,j)}$. Let $\mathbf{R}$ be the relation embeddings for all relations in $\mathcal{R}$, obtained using TransE. Each extracted edge $(v_i, r, v_j) \in \mathcal{E}$ is thus initialized in $\mathcal{G}'$ as $\mathbf{e}_{(i,j)} = \mathbf{r} \in \mathbf{R}$. However, due to edge sparsity, many edges do not have labeled relations and cannot be initialized this way. 

Meanwhile, despite PLMs' limitations in commonsense, they have shown some ability to encode commonsense knowledge \citep{davison2019commonsense, petroni2019language} and aid KG completion \citep{malaviya2019exploiting, bosselut2019comet, wang2020connecting}. Hence, we generate edge embeddings for all unlabeled edges by feeding each unlabeled edge into a GPT-2 \cite{radford2019language} based generator $f_\text{gen}(\cdot,\cdot)$. This is further explained in the ``Edge Embedding Generation" paragraph.

In summary, edge embeddings are computed in a hybrid way: \textbf{(1)} If there exists $r\in\mathcal{R}$ such that $(v_i, r, v_j)\in\mathcal{E}$, then $\mathbf{e}_{(i,j)} = \mathbf{r} \in \mathbf{R}$. \textbf{(2)} Otherwise, $\mathbf{e}_{(i,j)}=f_\text{adapt}(f_{\text{gen}}(v_i, v_j))$, where
$f_\text{adapt}(\cdot)$ is an MLP used to transform $f_{\text{gen}}(v_i, v_j)$ into the same space as $\mathbf{r}$.

\paragraph{Edge Embedding Generation.}
Inspired by recent work in PLM-based commonsense KG completion \citep{bosselut2019comet, malaviya2019exploiting, wang2020connecting}, we frame edge generation as text generation. First, for each extracted edge $(h, r, t) \in \mathcal{E}$, we first tokenize its node pair $(h, t)$ and relation label $r$. Let $\tilde{h}$, $\tilde{r}$, and $\tilde{t}$ be the respective token sequences of $h$, $r$, and $t$. Also, let $\$$ be the special separator token. Next, for each tokenized extracted edge, we train a GPT-2 model~\citep{radford2019language} to autoregressively generate the concatenated sequence $[\tilde{h},\$,\tilde{t},\$,\tilde{h},\tilde{r},\tilde{t}]$. 

During inference, we only have unlabeled edges $(v_i, v_j) \in \mathcal{E}'$, with no $r$. Thus, for each $(v_i, v_j)$, GPT-2 is given $s_{\text{input}} = [ \tilde{v_i},\$,\tilde{v_j},\$]$ and asked to generate the missing tokens $s_{\text{pred}} = [\tilde{v_i},\tilde{r},\tilde{v_j}]$.
Let $[x_1, x_2, ... , x_T] = [s_{\text{input}}, s_{\text{pred}}]$.
The edge embedding for $(v_i, v_j)$ is then computed as $f_\text{gen}(v_i,v_j)=\frac{1}{T}\sum_{i=1}^{T}\mathbf{h}_i$, where $\mathbf{h}_i$ is the GPT-2 hidden state for $x_i$. See Appendix \sectionref{sec:app_edge_gen} for more details.

Alternatively, we consider another edge generation approach proposed by \citet{wang2020connecting}. Here, $f_\text{gen}(\cdot,\cdot)$ is trained to generate a relational path connecting $v_i$ to $v_j$, then pool the path into an edge embedding. The rationale for this approach is that such paths have been shown to contain useful semantic information about the relation between $v_i$ and $v_j$ \citep{neelakantan2015compositional, das2017chains, wang2020connecting}.




\paragraph{Adjacency Matrix.}
Before edge generation, $\mathcal{G}'$ has binary adjacency matrix $\mathbf{A}^{\text{extract}}$, where $\mathbf{A}_{(i,j)} = 1 \Leftrightarrow \exists r, \text{s.t. } (v_i, r, v_j) \in \mathcal{E}$. After getting embeddings for all edges $(v_i, v_j) \in \mathcal{E}'$, $\mathbf{A}^{\text{extract}}$ becomes $\mathbf{A}^0$, a denser binary adjacency matrix in which $\mathbf{A}^0_{(i,j)} = 1 \Leftrightarrow (v_i, v_j) \in \mathcal{E}'$.

\subsection{Hybrid Graph Reasoning}
\label{subsec:graph_reasoning}

The procedure described in \sectionref{subsec:graph_construction} yields a hybrid graph, containing unweighted edges between all question-answer node pairs. Constructing this hybrid graph may improve edge recall, but does not address precision. Some edges in the initial hybrid graph may be irrelevant to the question-answer pair, either due to noisy edge extraction or generation. HGN is thus designed to downweight irrelevant edges by converting the unweighted graph into a weighted one, then learning to reweight all hybrid edges during reasoning (Figure~\ref{fig:hgn}).

\paragraph{Learnable Adjacency Matrix.}
Although $\mathbf{A}^0$ is a binary adjacency matrix, HGN populates it with learned edge attention weights and iteratively updates them over $L$ layers of reasoning. We denote the adjacency matrix at layer $\ell$ as $\mathbf{A}^\ell$, where $0 \leq \mathbf{A}^\ell_{(i,j)} \leq 1$. Updating $\mathbf{A}^\ell$ can be viewed as softly contextualizing the hybrid graph's structure with respect to $(q, a)$.

\paragraph{Edge-Weighted Message Passing.}
Following the general Graph Network (GN) formulation proposed by \citet{battaglia2018relational}, HGN's graph reasoning module consists of layer-wise node-to-edge ($v\rightarrow e$) and edge-to-node ($e\rightarrow v$) message passing functions.
However, we equip HGN with a modified version of GN's edge-to-node message passing function, in which each edge's weight is used to rescale information flow on that edge. Intuitively, an edge's weight signifies the edge's relevance for reasoning about the given task instance.
We also use text encoding $\mathbf{s}$ as global context throughout message passing. 

Formally, HGN's update rule at layer $\ell$ is:
\begin{equation}
\label{eq:gn}
\begin{split}
v\rightarrow e: \mathbf{h}^\ell_{(i,j)} &= f^\ell_{v\rightarrow e}\left(\left[\mathbf{h}^{\ell-1}_i,\mathbf{h}^{\ell-1}_j,\mathbf{h}^{\ell-1}_{(i,j)},\mathbf{s}\right]\right);\\
w^\ell_{(i,j)} &= f^\ell_w\left(\left[\mathbf{h}^{\ell-1}_{(i,j)},\mathbf{s}\right]\right);\\
\mathbf{A}^\ell_{(i,j)} &= \frac{e^{w^\ell_{(i,j)}}}{\sum_{(s,t)\in \mathcal{E}'}e^{w^\ell_{(s,t)}}},\\
e\rightarrow v:
\mathbf{u}^\ell_{(i,j)} &= f^\ell_u\left(\left[\mathbf{h}^{\ell-1}_i,\mathbf{h}^\ell_{(i,j)}\right]\right);\\
\mathbf{h}^\ell_j &= f^\ell_{e\rightarrow v}\left(\sum\nolimits_{i\in\mathcal{N}_j}\mathbf{A}^\ell_{(i,j)}\mathbf{u}^\ell_{(i,j)}\right).
\end{split}
\end{equation}
$N_j$ is the set of $v_j$'s incoming neighbors; $f^\ell_{v\rightarrow e}$, $f^\ell_w$, $f^\ell_u$ and $f^\ell_{e\rightarrow v}$ are MLPs; $\mathbf{h}_{(i,j)}^0=\mathbf{e}_{(i,j)}$ is the initial embedding for edge $(v_i, v_j)$; and $\mathbf{h}_i^0=\mathbf{v}_i$ is the initial embedding for node $v_i$.

In node-to-edge message passing, the embedding of each edge $(v_i,v_j) \in \mathcal{E}'$ is updated as $\mathbf{h}^\ell_{(i,j)}$, a function of $(v_i,v_j)$'s constituent nodes and the given context $\mathbf{s}$. Through $\mathbf{s}$, the hybrid graph is strongly contextualized with respect to $(q, a)$.
Then, $\mathbf{h}^\ell_{(i,j)}$ is used to compute edge score $w^\ell_{(i,j)}$, which measures the edge's relevance to $\mathbf{s}$.
Each edge score is globally normalized across all edges in the graph to produce edge attention weight $\mathbf{A}^\ell_{(i,j)}$, so that low-scoring edges are softly pruned by receiving close-to-zero weight.

We use global edge attention (i.e., normalizing across $\mathcal{E}'$) instead of local edge attention (i.e., normalizing across $N_j$) because local edge attention assumes at least one edge in $N_j$ is relevant, which may not be true. For example, given an irrelevant or incorrectly grounded concept, none of its edges will be helpful, and so all nodes in its neighborhood should be excluded from influencing the reasoning process. To demonstrate the advantage of global edge attention, we empirically compare our default HGN architecture to an HGN variant based on Graph Attention Network (GAT)~\citep{velivckovic2017graph}, which uses local edge attention, in our experiments.

In edge-to-node message passing, the embedding of each node $v_j \in \mathcal{V}'$ is updated as $\mathbf{h}^\ell_{j}$, a function of $v_j$'s neighboring edges. For each edge neighbor, edge weight $\mathbf{A}_{(i,j)}^\ell$ is used to rescale the edge's influence on $v_j$'s embedding update.

\subsection{Learning Objective}
\label{subsec:learning_objective}
\paragraph{Task Loss.}
After $L$ layers of message passing, we obtain node embeddings $\{\mathbf{h}^L_i\mid i: v_i\in \mathcal{V}'\}$ and edge embeddings $\{\mathbf{h}^L_{(i,j)}\mid (i, j): (v_i, v_j)\in  \mathcal{E}'\}$.
Node embeddings are aggregated into $\mathbf{v}_\text{agg}$ via attentive pooling with $\mathbf{s}$ as the query vector.
Edge embeddings are aggregated into $\mathbf{e}_\text{agg}$ via edge-weighted sum pooling.
The final graph encoding is then given as $\mathbf{g}=[\mathbf{v}_\text{agg}, \mathbf{e}_\text{agg}]$.
The probability of $a$ being the answer to $q$ is calculated as $\hat{\rho}(q,a) \propto \exp(\rho(q,a))$, where $\rho(q,a)=f_\text{score}([\mathbf{s}, \mathbf{g}];\boldsymbol{\uptheta}_\text{score})$.
We use cross-entropy loss for the QA classification task, so the loss for each $(q,a)$ with label $y$ is:
\begin{equation}
\mathcal{L}_\text{task}\left(\hat{\rho}(q,a;\boldsymbol{\uptheta})),y\right)= -y\log\hat{\rho}(q,a;\boldsymbol{\uptheta}).
\end{equation}

\paragraph{Entropy Regularization.} 
To encourage the model to be decisive during edge reweighting, we use a regularization term to penalize non-discriminative edge weights. In an extreme case, a blind model will assign the same weight to all edges, degenerating $\mathcal{G}'$ into an unweighted graph.
This is a failure mode, since $\mathcal{G}'$ is likely to contain mostly irrelevant edges, and we want the model to focus on the helpful edges.
Therefore, via $\mathcal{L}_\text{edge}$, we train the model to minimize the entropy of the edge weight distribution (i.e., make the distribution more skewed), in order to maximize the informativeness of the predicted edge weights.
Lower entropy means the model has higher certainty about edges' relevance to the given task instance, such that the model will discriminatively judge some edges as being much more relevant than others.
$\mathcal{L}_\text{edge}$ is computed as:
\begin{equation}
\begin{split}
\mathcal{L}_\text{edge}(\mathbf{A}^L(q,a))= - \hspace{-0mm} \sum_{(i,j):(v_i,v_j)\in \mathcal{E}'} \hspace{-0mm} \mathbf{A}^L_{(i,j)} \log \mathbf{A}^L_{(i,j)}.
\end{split}
\end{equation}

\paragraph{Joint Learning.}
We jointly optimize $\mathcal{L}_\text{task}$ and $\mathcal{L}_\text{edge}$, so graph reasoning and structure can be jointly learned.
The full learning objective is:
\begin{small}
\begin{equation}
\begin{split}
\label{eq:obj}
\begin{aligned}
\mathcal{L}(\boldsymbol{\uptheta}) = \hspace{-0mm} \sum_{(q,a,y)\sim X_\text{train}} &\Big{[}\mathcal{L}_\text{task}\left(\hat{\rho}(q,a)),y\right)+ \beta\cdot \mathcal{L}_\text{edge}(\mathbf{A}^L(q,a)) \Big{]},
\end{aligned}
\end{split}
\end{equation}
\end{small}
where $\boldsymbol{\uptheta}=\{\boldsymbol{\uptheta}_\text{text}, \boldsymbol{\uptheta}_\text{graph}, \boldsymbol{\uptheta}_\text{score}\}$ is the set of all learnable parameters, and $X_{\text{train}}$ is the training set.
We train our model end-to-end by minimizing $\mathcal{L}(\boldsymbol{\uptheta})$ with the RAdam \citep{liu2019radam} optimizer. 

%% file: sections/4_experiments.tex
\section{Experiments}
\label{sec:experiments}

\begin{table*}[t]
\centering
\scalebox{0.7}{
\begin{tabular}{lcccccc}
    \toprule  
    \multirow{2}{*}{\textbf{Methods}}&
    \multicolumn{2}{c}{ \textbf{BERT-Base}}&\multicolumn{2}{c}{ \textbf{BERT-Large}}&\multicolumn{2}{c}{ \textbf{RoBERTa}}\\
    \cmidrule(lr){2-3} \cmidrule(lr){4-5} \cmidrule(lr){6-7}
     & 60\% Train & 100\% Train & 60\% Train & 100\% Train & 60\% Train & 100\% Train \\
    \midrule  
    {LM Finetuning$^*$}  & 52.06~($\pm$0.72)  &  53.47~($\pm$0.87)  &  52.30~($\pm$0.16) & 55.39~($\pm$0.40) & 65.56~($\pm$0.76) & 68.69~($\pm$0.56) \\
    \midrule
    {RN$^*$~\citep{santoro2017simple}}  &  54.43~($\pm$0.10)  &  56.20~($\pm$0.45)  &  54.23~($\pm$0.28) & 58.46~($\pm$0.71) & 66.16~($\pm$0.28) & 70.08~($\pm$0.21) \\
    {RN + Link Prediction$^*$}  & - & - & 53.96~($\pm$0.56) &  56.02~($\pm$0.55) & 66.29($~\pm$0.29) & 69.33~($\pm$0.98)\\
    {RGCN$^*$~\citep{Schlichtkrull2017ModelingRD}} &  52.20~($\pm$0.31)  &  54.50~($\pm$0.56)  &  54.71~($\pm$0.37) & 57.13~($\pm$0.36) & 68.33~($\pm$0.85) & 68.41~($\pm$0.66) \\
    {GAT~\citep{velivckovic2017graph}} &  53.05~($\pm$0.37)  &  56.51~($\pm$0.74)  &  55.80~($\pm$0.53) & 58.18~($\pm$1.07) & 69.63~($\pm$0.42) & 71.20~($\pm$0.72) \\
    {GN~\citep{battaglia2018relational}}  & 53.67~($\pm$0.45) & 55.65~($\pm$0.51) & 54.78~($\pm$0.61) &  57.81~($\pm$0.67) & 68.78~($\pm$0.67) & 71.12~($\pm$0.45)\\
    {GconAttn$^*$~\citep{Wang2018ImprovingNL}}  & 51.36~($\pm$0.98) & 54.41~($\pm$0.50) & 54.96~($\pm$0.69) &  56.94~($\pm$0.77) & 68.09~($\pm$0.63) & 69.88~($\pm$0.47)\\
    {KagNet$^*$~\citep{lin2019kagnet}}  &  -  &  56.19  &  - & 57.16 & - & - \\
    {MHGRN$^*$~\citep{feng2020scalable}}  & 54.12~($\pm$0.49) & 56.23~($\pm$0.82) & 56.76~($\pm$0.21) &  59.85~($\pm$0.56) & 68.84~($\pm$1.06) & 71.11~($\pm$0.81)\\
    {PathGenerator$^*$~\citep{wang2020connecting}}  & 54.44~($\pm$0.42) & 56.99~($\pm$0.41) & 57.53~($\pm$0.19) &  59.07~($\pm$0.30) & 69.46~($\pm$0.23) & 72.68~($\pm$0.42)\\
    \midrule
    {HGN (w/ PathGen edges)} & 55.68~($\pm$0.29) & 57.77~($\pm$0.39) & 58.19~($\pm$0.27) & 60.89~($\pm$0.19) & 70.95~($\pm$0.21) & 73.41~($\pm$0.31) \\
    {HGN (w/ RelGen edges)} & \textbf{55.72}~($\pm$0.32) & \textbf{58.01}~($\pm$0.29) & \textbf{58.19}~($\pm$0.11) & \textbf{61.11}~($\pm$0.21) & \textbf{71.10}~($\pm$0.11) & \textbf{73.64}~($\pm$0.30) \\
    \bottomrule 
\end{tabular}
}
\caption{\textbf{Accuracy on CommonsenseQA inhouse test set.} Both our model variants significantly improve over all baselines. We use the same inhouse split as \citet{lin2019kagnet}. For  baselines with $^*$, we use the reported numbers from \citet{feng2020scalable} and \citet{wang2020connecting} if available. Mean and standard deviation of four seeds are presented for all models except KagNet.}
\label{tab:csqa_main}
\end{table*}

\subsection{Experimental Setup}
\label{subsec:exp:setup}
We evaluate our proposed model on four multiple-choice commonsense QA datasets: \textbf{CommonsenseQA}~\citep{talmor2019commonsenseqa}, \textbf{CODAH} ~\citep{chen2019codah}, \textbf{OpenBookQA}~\citep{mihaylov2018can} and \textbf{QASC}~\citep{khot2020qasc} (details in Appendix~\sectionref{sec:app_datasets}). We use ConceptNet~\citep{speer2017conceptnet}, a commonsensense knowledge graph, as $\mathcal{G}$. For text encoder $f_\text{text}$, we experiment with BERT-Base, BERT-Large~\citep{devlin2018bert} and RoBERTa(-Large)~\citep{liu2019roberta} to validate our model's effectiveness over different text encoders. For OpenbookQA and QASC, retrieving related facts from the provided corpus plays an important role in boosting the model's performance.
Therefore, we build our graph reasoning model on top of retrieval-augmented methods on the leaderboard: ``AristoRoBERTa''\footnote{ \url{https://leaderboard.allenai.org/open_book_qa/submission/blcp1tu91i4gm0vf484g}} for OpenBookQA and ``RoBERTa (2-step IR)''\footnote{\url{https://leaderboard.allenai.org/qasc/submission/bolaun0ghifmkohgvhr0}} for QASC. In this way, we can study if strong retrieval-augmented methods can still benefit from KG knowledge and our HGN framework.

\subsection{Compared Methods}
\label{subsec:exp:compared_methods}
We compare our model with a series of KG-augmented methods and different graph encoders: 

\paragraph{Models Using Extracted Facts.} 
We consider seven models that only use extracted facts.
\textbf{RN}~\citep{santoro2017simple} builds the graph with the same node set as our method but extracted edges only. The graph vector is calculated as $\mathbf{g}=\text{Pool}(\{\text{MLP}([\mathbf{v}_i,\mathbf{e}_{(i,j)},\mathbf{v}_j])\mid (v_i,v_j)\in\mathcal{E}'\})$.
\textbf{GN}~\citep{battaglia2018relational} presents a general formulation of GNNs. We instantiate it with the layerwise propagation rule defined in Equation~\ref{eq:gn}. It differs from our HGN in that: (1) it only considers extracted edges; (2) all edge weights are fixed to 1.
\textbf{MHGRN}~\citep{feng2020scalable} generalizes GNNs with multi-hop message passing.
\textbf{GAT}~\citep{velivckovic2017graph} adopts attention mechanism to reweight edges locally in each node's neighborhood. We implement it by replacing the graph edge attention with local edge attention and only considering $\mathcal{L}_\text{task}$ during training.
\textbf{RGCN}~\citep{schlichtkrull2018modeling} extends Graph Convolutional Networks (GCNs)~\citep{kipf2016semi} with relation-specific transition matrices during message passing. It operates on the same graph as RN. The graph vector is calculated as $\mathbf{g}=\text{Pool}(\{\mathbf{h}^L_i\mid v_i\in V\})$.
\textbf{GconAttn}~\citep{wang2019improving} softly aligns the nodes in question and answer and do pooling over all matching nodes to get $\mathbf{g}$.
\textbf{KagNet}~\citep{lin2019kagnet} uses an LSTM to encode relational paths between question and answer concepts and pool over the path embeddings for graph encoding.

\paragraph{Models Using Extracted and Generated Facts.} 
We consider two models that use both extracted facts and generated facts.
\textbf{RN + Link Prediction} differs from RN by only considering the generated relation (predicted using TransE~\citep{bordes2013translating}) between question and answer concepts.
\textbf{PathGenerator}~\citep{wang2020connecting} learns a path generator from paths collected through random walks on the KG. The learned generator is used to generate paths  connecting question and answer concepts. $\mathbf{g}$ is calculated as the concatenation of the pooled vector over the generated paths and the pooled vector over the extracted paths.

\paragraph{Our Model's Variants.} 
As described in \sectionref{subsec:graph_construction}, the edge embedding can be computed either as a relation embedding or a path embedding. We name these two variants as \textbf{HGN (w/ RelGen edges)} and \textbf{HGN (w/ PathGen edges)} respectively.

\begin{table}[t]
\centering
\scalebox{0.7}{
\begin{tabular}{lcccccc}
    \toprule  
    {\textbf{Methods}} & \textbf{Single} & \textbf{Ensemble} \\
    \midrule  
    {ALBERT+DESC-KCR~\citep{xu2020fusing}} & 80.7 & 83.3  \\
    {ALBERT+KD} & 80.3 & 80.9 \\
    {ALBERT+KCR} & 79.5 & - \\
    {Unified QA~\citep{khashabi2020unifiedqa}} & 79.1 & -  \\
    {ALBERT+KRD} & 78.4 & -  \\
    {T5-3B~\citep{raffel2019exploring}} & 78.1 & -  \\
    {\textbf{ALBERT+HGN (w/ RelGen edges)}} & \textbf{77.3} & \textbf{80.0}  \\
    {TeGBERT} & 76.8 & - \\
    {ALBERT+PathGenerator~\citep{wang2020connecting}} & 75.6 & 78.2  \\
    {ALBERT~\citep{lan2019albert}} & - & 76.5  \\
    \bottomrule
\end{tabular}
}
\caption{\textbf{Leaderboard of CommonsenseQA.} HGN ranks first among comparable systems, especially achieving remarkable improvement over PathGenerator \citep{wang2020connecting}.}
\label{tab:csqa_leaderboard}
\end{table}

\begin{table}[t]
\centering
\scalebox{0.7}{
\begin{tabular}{lcccccc}
    \toprule  
    {\textbf{Methods}}&
    \textbf{BERT-Large} & \textbf{RoBERTa}\\
    \midrule  
    {LM Finetuning}  & 65.74 & 83.14 \\
    \midrule
    {RN~\citep{santoro2017simple}}  & 64.59 & 82.45 \\
    {RGCN~\citep{Schlichtkrull2017ModelingRD}}  & 65.56 & 82.42 \\
    {GAT~\citep{velivckovic2017graph}}  & 65.88 & 82.78 \\
    {GN~\citep{battaglia2018relational}}  & 65.52 & 82.06 \\
    {GconAttn~\citep{Wang2018ImprovingNL}}  & 65.17 & 82.35 \\
    {MHGRN~\citep{feng2020scalable}}  & 65.92 & 83.07 \\
    {PathGenerator~\citep{wang2020connecting}}  & 64.67 & 82.27 \\
    \midrule
    {HGN (w/ PathGen edges)} & 66.21 & \textbf{84.32} \\
    {HGN (w/ RelGen edges)}  & \textbf{66.75} & 84.08 \\
    \bottomrule
\end{tabular}
}
\caption{\textbf{Test accuracy on CODAH.} Both our model variants consistently improve over all baselines. We use the official split for 5-fold cross validation. Mean accuracy on 5 folds are presented.}
\label{tab:codah_main}
\end{table}

\begin{table}[ht]
\scalebox{0.59}{
\begin{tabular}{lcccccc}
    \toprule  
    {\textbf{Datasets}} & \textbf{OpenBookQA} & \textbf{QASC}\\
    {\textbf{Base Models}} & \textbf{AristoRoBERTa} & \textbf{RoBERTa (2-step IR)}\\
    \midrule  
    {LM Finetuning$^*$} & 77.40~($\pm$1.64) & 73.34~($\pm$0.71) \\
    \midrule
    {RN$^*$~\citep{santoro2017simple}} & 78.05~($\pm$0.77) & 72.77~($\pm$1.50) \\
    {RN + Link Prediction$^*$} & 77.25~($\pm$1.11) & - \\
    {RGCN$^*$~\citep{Schlichtkrull2017ModelingRD}} & 74.60~($\pm$2.53) & 72.23~($\pm$1.36) \\
    {GAT~\citep{velivckovic2017graph}} & 78.20~($\pm$1.22) & 72.61~($\pm$0.93) \\
    {GN~\citep{battaglia2018relational}} & 77.25~($\pm$0.91) & 72.53~($\pm$0.70)   \\
    {GconAttn$^*$~\citep{Wang2018ImprovingNL}} & 71.80~($\pm$1.21) & 72.72~($\pm$1.66)   \\
    {MHGRN~\citep{feng2020scalable}} & 77.75~($\pm$0.38) & 73.24~($\pm$0.45) \\
    {PathGenerator$^*$~\citep{wang2020connecting}} & 79.15~($\pm$0.78) & 72.96~($\pm$0.68) \\
    \midrule
    {HGN (w/ PathGen edges)} & 80.05~($\pm$0.54)  &  74.10~($\pm$0.42) \\
    {HGN (w/ RelGen edges)}  & \textbf{80.15}~($\pm$0.38)  & \textbf{74.27}~($\pm$0.31) \\
    \bottomrule
\end{tabular}
}
\caption{\textbf{Test accuracy on OpenBookQA and QASC with retrieval-augmented methods as base models.} Both our model variants greatly improve over all baselines except HGN (w/ PathGen edges) over MHGRN. For OpenbookQA baselines with $^*$, we use reported numbers from \citet{wang2020connecting}. Mean and standard deviation of four seeds are presented.}
\label{tab:obqa_main}
\end{table}

\begin{table}[t]
\centering
\scalebox{0.7}{
\begin{tabular}{lcccccc}
    \toprule  
    {\textbf{Methods}} & \textbf{Text Encoder} & \textbf{Test Acc} \\
    \midrule  
    {UnifiedQA~\citep{khashabi2020unifiedqa}} & T5-11B & 87.2 \\
    {T5-11B + KB~} & T5-11B & 85.4  \\
    {T5-3B~\citep{raffel2019exploring}} & T5-3B  & 83.2  \\
    {PathGenerator}~\citep{wang2020connecting} &  ALBERT &  81.8 \\
    {\textbf{HGN (w/ RelGen edges)}} &  \textbf{AristoRoBERTa} &  \textbf{81.4} \\
    {AristoRoBERTa + KB} &  AristoRoBERTa &  81.0 \\
    {MHGRN}~\citep{feng2020scalable} &  AristoRoBERTa &  80.6 \\
    {PathGenerator}~\citep{wang2020connecting} &  AristoRoBERTa &  80.2 \\
    {KF + SIR}~\citep{banerjee2020knowledge} &  RoBERTa &  80.2 \\
    {AristoRoBERTa} &  AristoRoBERTa &  80.2 \\
    \bottomrule
\end{tabular}
}
\caption{\textbf{Leaderboard of OpenBookQA.} Our HGN ranks first among all submissions using AristoRoBERTa as the text encoder.}
\label{tab:obqa_leaderboard}
\end{table}

\begin{figure*}[t]
\centering
\subfloat[][CommonsenseQA (RoBERTa).]{\includegraphics[width=0.32\textwidth]{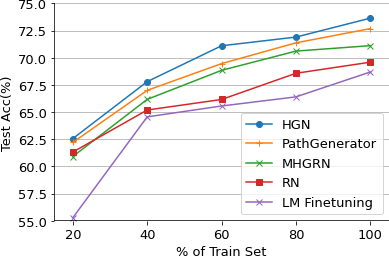}}
\subfloat[][OpenBookQA (AristoRoBERTa).]{\includegraphics[width=0.32\textwidth]{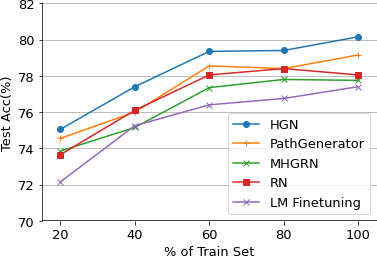}}
\subfloat[][Ablations on model variants.]{\includegraphics[width=0.36\textwidth]{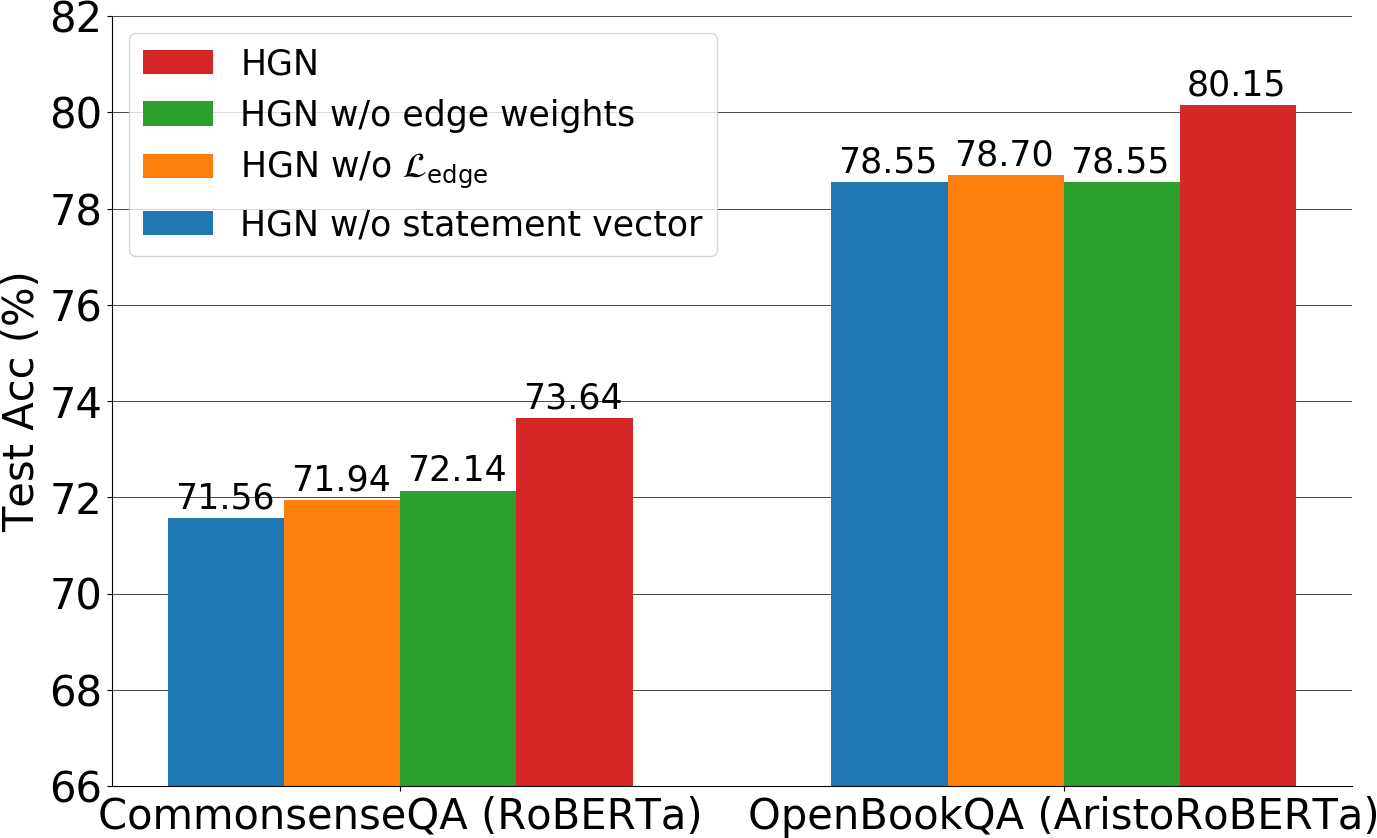}}
\caption{\textbf{Low-resource and ablation studies.} (a)(b) Performance of HGN and baseline models with different amounts of training data; (c) Performance of different model variants.
}
\label{fig:ablation}
\end{figure*}



\subsection{Results}
\label{sec:results}

\paragraph{Performance Comparisons.}
Tables~\ref{tab:csqa_main}, ~\ref{tab:codah_main}, ~\ref{tab:obqa_main} show performance comparisons between our models and baseline models on CommonsenseQA, CODAH, OpenBookQA and QASC.
We clearly find that models with stronger text encoders perform better (i.e. RoBERTa $>$ BERT-Large $>$ BERT-Base).
For all text encoders, our HGN shows consistent improvement over baseline models on all datasets. The improvement over all baselines are tested to be statistically significant under most settings, demonstrating the effectiveness of HGN both with and without retrieved evidence.

We also submit our best model to leaderboards for CommonsenseQA and OpenBookQA.
For CommonsenseQA (Table~\ref{tab:csqa_leaderboard}), our HGN ranks first among comparable approaches and shows remarkable improvement over PathGenerator~\citep{wang2020connecting} and the LM Finetuning approach (ALBERT~\citep{lan2019albert}).
Higher-ranking models either use stronger text encoders or leverage additional data resources.
Specifically, UnifiedQA~\citep{khashabi2020unifiedqa} and T5-3B~\citep{raffel2019exploring} are based on T5.
They have 11B and 3B parameters respectively, making them impractical to be finetuned in an academic setting.
ALBERT+DESC-KCR~\citep{xu2020fusing} and ALBERT+KD additionally use concept definitions from dictionaries.
ALBERT+DESC-KCR and ALBERT+KCR leverage ``question\_concept'' annotations, which are used during the construction of the CommmonsenseQA dataset and allow the model to learn shortcuts that don't generalize to other datasets.
ALBERT+KRD retrieve sentences from OMCS corpus~\citep{liu2004conceptnet} as input.
These methods are therefore not comparable with our model.
For OpenBookQA (Table~\ref{tab:obqa_leaderboard}), our model ranks first among all models using AristoRoBERTa as the text encoder.


\paragraph{Training with Less Labeled Data.}
Figure~\ref{fig:ablation} (a)(b) show the results of our model and baselines when trained with different portions of the training data on CommonsenseQA and OpenBookQA.
Our model gets better test accuracy under all settings.
On CommonsenseQA without retrieved evidence, the improvement over the knowledge-agnostic baseline (LM Finetuning) is generally more significant with less training data, which suggests that incorporating external knowledge is helpful in the low-resource setting.

\paragraph{Study on More Model Variants.}
To better understand the model design, we experiment with three variants of HGN (w/ RelGen edges) on CommonsenseQA and OpenBookQA.
\textbf{HGN w/o statement vector} doesn't consider $\mathbf{s}$ in Equation~\ref{eq:gn}, which isolates the graph encoder from the text encoder.
\textbf{HGN w/o $ \mathcal{L}_\text{edge}$} does not consider the entropy regularization term and thus does not penalize non-discriminative edge weights.
\textbf{HGN w/o edge weights} reasons over an unweighted graph with hybrid features, which means edge weights are all fixed to 1 during training.
Figure~\ref{fig:ablation} (c) shows the results of the ablation study.
``HGN'' outperforms ``HGN w/o $ \mathcal{L}_\text{edge}$'', suggesting the usefulness of our proposed entropy regularization.
Comparing ``HGN w/o statement vector'' with ``HGN'', we find that accessing context information is also important for graph reasoning, which means information propagation and edge weight prediction should be conducted in a context-aware manner. HGN also improves over ``HGN (w/o edge weights)'', indicating the effectiveness of conducting context-dependent pruning.

\begin{table}[t]
\scalebox{0.7}{
\begin{tabular}{lcccccc}
    \toprule  
    {\textbf{Contextualized Graph}}&
    \textbf{GN} ($\mathbf{A}^\text{extract}$) & \textbf{HGN} ($\mathbf{A}^{L}$)\\
    \midrule  
    {Number of Edges}  & 3.65~($\pm$2.73) & 4.38~($\pm$3.24) \\
    {Number of Valid Edges}  & 2.67~($\pm$1.95) & 3.15~($\pm$1.98) \\
    {Percentage of Valid Edges}  & 71.64\% & 78.51\% \\
    {Average Helpfulness Score of Edges}  & 0.90~($\pm$0.50) &1.16~($\pm$0.51)\\
    {Prune Rate}  & - & 22.84\% \\
    \bottomrule
\end{tabular}
}
\caption{\textbf{User studies on learned graph structures.} 30 pairs of contextualized graphs output by GN and HGN are evaluated by 5 annotators.}
\label{tab:user_study}
\end{table}

\subsection{User Study on Learned Structures}
To assess HGN's ability to refine graph structure, we compare the graph structure before and after being processed by HGN. Specifically, we sample 30 questions with its answer from CommonsenseQA's development set and ask 5 human annotators to evaluate the graph output by GN (with adjacency matrix $\mathbf{A}^\text{extract}$ and extracted facts only) and by HGN (with adjacency matrix $\mathbf{A}^L$). We manually binarize $\mathbf{A}^L$ by removing edges with weight lower than $0.01$.

Given a graph, for each edge (fact), annotators are asked to rate its \textbf{validness} and \textbf{helpfulness}. The validness score is rated as a binary value in a context-agnostic way: 0 (the fact does not make sense), 1 (the fact is generally true). The helpfulness score measures if the fact is helpful for solving the question and is rated on a 0 to 2 scale: 0 (the fact is unrelated to the question and answer), 1 (the fact is related but doesn't directly lead to the answer), 2 (the fact directly leads to the answer).
Note that the percentage of valid edges can be understood as the precision of graph edges.
For a given instance, the number of valid edges is proportional to the recall of the edges.
We also include another metric named ``prune rate'' calculated as: $1-\frac{\text{\# edges in binarized } \mathbf{A}^L}{\text{\# edges in } \mathbf{A}^0}$, which measures the portion of edges assigned very low weights (softly pruned) during training and is only applicable to HGN.

The mean ratings for 30 pairs of (GN, HGN) graphs by 5 annotators are reported in Table \ref{tab:user_study}. 
The Fleiss' Kappa~\citep{fleiss1971measuring} is 0.51 (moderate agreement) for validness and 0.36 (fair agreement) for helpfulness. The graph refined by HGN has both more edges and denser valid edges compared to the extracted one. The refined graph also achieves a higher average helpfulness score. These all indicate that our HGN learns a superior graph structure with more helpful edges and fewer noisy edges, which improves over previous works that rely on extracted and static graphs. Detailed cases can be found in Appendix \sectionref{subsec:Appendix:Case Study}.

%% file: sections/5_related.tex
\section{Related Work}
\label{sec:related}

\paragraph{Commonsense QA.}
Commonsense QA is challenging because the required commonsense knowledge is seldom given in the question-answer context or encoded in the PLM's parameters. Thus, many works obtain this knowledge from external sources (e.g., KGs, corpora). While \citet{lv2020graph} show that KGs and corpora can provide complementary knowledge, our paper focuses on improving the use of KG knowledge. KG knowledge can be acquired in different ways, either from KG-extracted edges \cite{lin2019kagnet, ma2019towards, feng2020scalable, yasunaga2021qa}, PLM-generated edges \cite{bosselut2019dynamic}, or both \cite{wang2020connecting}. KG-augmented models mainly differ in how they encode KG knowledge, using message passing \cite{schlichtkrull2018modeling, feng2020scalable} or edge/path aggregation \cite{lin2019kagnet, bosselut2019dynamic, ma2019towards, wang2020connecting}.
The most relevant work to ours is \citet{wang2020connecting}.
The main difference is that they coarsely combine extracted and generated knowledge via late fusion, while HGN encodes both types of knowledge within a unified graph.
Besides, they use RN to pool over a set of paths for graph encoding, while HGN reasons over the graph via message passing and edge reweighting.



\vspace{-0.0cm}
\paragraph{Graph Structure Learning.}
Instead of assuming a fixed graph structure, a number of graph models learn the graph structure with respect to the downstream task. 
Some models learn to discretely select edges for the graph (i.e., hard pruning).
\citet{kipf2018neural} and \citet{franceschi2019learning} sample the graph structure from a predicted probabilistic distribution with differentiable approximations.
\citet{norcliffe2018learning} calculate the relatedness between any pair of nodes and only keep the top-$k$ strongest connections for each node to construct the edge set.
\citet{sun2019pullnet} start with a small graph and iteratively expand it with retrieving operations.
Others learn to reweight edges in a fully connected graph (i.e., soft pruning).
\citet{jiang2019semi} and \citet{yu2019graph} propose heuristics for regularizing edge weights.
\citet{hu2019language} use the question embedding to help predict edge weights.
Unlike other edge reweighting models, HGN operates over a hybrid graph of both extracted and generated edges, while updating edge weights with respect to node, edge, and text features.

%% file: sections/6_conclusion.tex
\section{Conclusion}
\label{sec:conclusion}

In this paper, we propose HGN, a KG-augmented model for CSR. To address KG edge sparsity and noisy edge extraction/generation, HGN learns to jointly contextualize extracted and generated knowledge by reasoning over both within a unified graph structure. We justify HGN's design by showing that HGN improves performance on various CSR benchmarks and user studies. In future work, we plan to increase the graph's relation expressiveness by incorporating open relations, plus make the edge extraction/generation process more dependent on the reasoning context.

%% file: sections/7_acknowledgements.tex
\section*{Acknowledgments}
This research is supported in part by the Office of the Director of National Intelligence (ODNI), Intelligence Advanced Research Projects Activity (IARPA), via Contract No. 2019-19051600007, the DARPA MCS program under Contract No. N660011924033 with the United States Office Of Naval Research, the Defense Advanced Research Projects Agency with award W911NF-19-20271, and NSF SMA 18-29268. The views and conclusions contained herein are those of the authors and should not be interpreted as necessarily representing the official policies, either expressed or implied, of ODNI, IARPA, or the U.S. Government. We would like to thank all the collaborators in USC INK research lab for their constructive feedback on the work.
We would also like to thank the anonymous reviewers for their valuable comments.

%% file: sections/8_appendix.tex
\section{Implementation Details of Edge Embedding Generator (RelGen)}
\label{sec:app_edge_gen}
Here, we give a more detailed explanation of the PLM-based edge embedding generator $f_\text{gen}$, introduced in the ``Edge Embedding Generation" paragraph of \sectionref{subsec:graph_construction}.

To implement $f_\text{gen}$, we adopt GPT-2~\citep{radford2019language}, which is pretrained on 
large corpora and achieves great success on a wide range of tasks involving sentence generation, as a generator to generalize the facts from the knowledge graph. We first convert each fact $(h,r,t)\in\mathcal{E}$ into a word sequence with a ``prompt-generation'' format: $\left[ \tilde{h},\$,\tilde{t},\$,\tilde{h},\tilde{r},\tilde{t}\right]$, where $\tilde{h}, \tilde{r}, \tilde{t}$ are the word sequence of $h,r,t$ respectively, $\$$ denotes the delimiter token used by GPT-2, and $[\cdot,\cdot]$ denotes word sequence concatenation.
We adopt this format because $\left[\tilde{h},\tilde{r},\tilde{t}\right]$ is similar to a natural language fact.
Generating facts in a natural format helps induce commonsense knowledge stored in GPT-2 \citep{bosselut2019comet}.
We denote the synthetic sentence as $s_{(h,r,t)}=\left[x_1^{(h,r,t)},\ldots,x_{n_{(h,r,t)}}^{(h,r,t)}\right]$ 
and finetune GPT-2 on all synthetic sentences created from $\mathcal{E}$ with the language modeling objective:
\begin{equation*}
\begin{split}
&\mathcal{L}_\text{gen}(\mathcal{E})=\\
&\sum_{(h,r,t)\in\mathcal{E}}
\hspace{-1mm}
\sum_{i=1}^{n_{(h,r,t)}}
\hspace{-2mm}
\log P\left(x_i^{(h,r,t)}\mid x_1^{(h,r,t)},\ldots,x_{i-1}^{(h,r,t)}\right).
\end{split}
\end{equation*}
After that, given any two concepts $(v_i, v_j)$, we build a prompt as $\left[ \tilde{v_i},\$,\tilde{v_j},\$\right]$ and let the model to generate the following word sequence. We denote the whole sentence (both prompt and generation) as $s_{(v_i, v_j)}$, and the hidden states of each word during generation as $\mathbf{h}_1,\ldots,\mathbf{h}_T$ where $T$ is the sentence length. We average hidden states of all words in the sentence to get the relational feature: $f_\text{gen}(v_i,v_j)=\frac{1}{T}\sum_{i=1}^{T}\mathbf{h}_i$.

\section{Details of Datasets}
Below are descriptions of the four datasets used for the experiments presented in \sectionref{sec:experiments}.
\label{sec:app_datasets}
\paragraph{CommonsenseQA}
\citep{talmor2019commonsenseqa} is a multiple-choice QA dataset targeting commonsense. It's constructed based on the knowledge in ConceptNet. Since the test set of the official split (9741/1221/1140 for OFtrain/OFdev/OFtest) is not publicly available, we compare our models with baseline models on the inhouse split (8500/1221/1241 for IHtrain/IHdev/IHtest)\footnote{\scriptsize \url{https://github.com/INK-USC/MHGRN/blob/master/data/csqa/inhouse_split_qids.txt}} used by previous works~\citep{lin2019kagnet, feng2020scalable, wang2020connecting}. 

\paragraph{CODAH}
\citep{chen2019codah} contains 2801 sentence completion questions testing commonsense reasoning skills. We perform 5-fold cross validation using the official split.

\paragraph{OpenBookQA}
\citep{mihaylov2018can} is a multiple-choice QA dataset modeled after open-book exams. Besides 5957 elementary-level science questions (4957/500/500 for train/dev/test), it also provides an open book with 1326 core science facts. Solving the dataset requires combining facts from open book with commonsense knowledge.

\paragraph{QASC}
\citep{khot2020qasc} is a QA dataset with questions about grade-school science. It has 9980 8-way multiple-choice questions (8134/926/920 train/dev/test), and comes with a corpus of 17M sentences. Since the official test set does not have labels, we create an in-house test split by moving a randomly sampled set of 920 questions from the training set to the test set. Solving questions in QASC requires retrieving facts from the corpus and composing them to produce an answer.



\section{Case Study}
\label{subsec:Appendix:Case Study}
\begin{figure*}[ht]
\centering
\subfloat[\textbf{Case I: Unrelated extracted facts are filtered out.}]{%
  \includegraphics[scale=0.45]{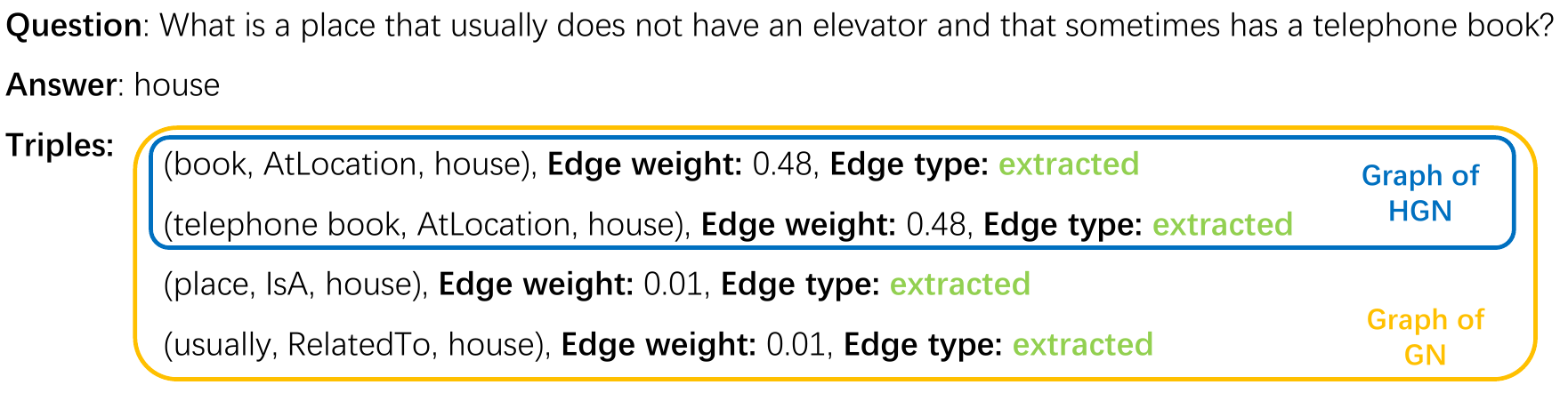}%
}

\subfloat[\textbf{Case II: Helpful generated facts are incorporated.}]{%
  \includegraphics[scale=0.45]{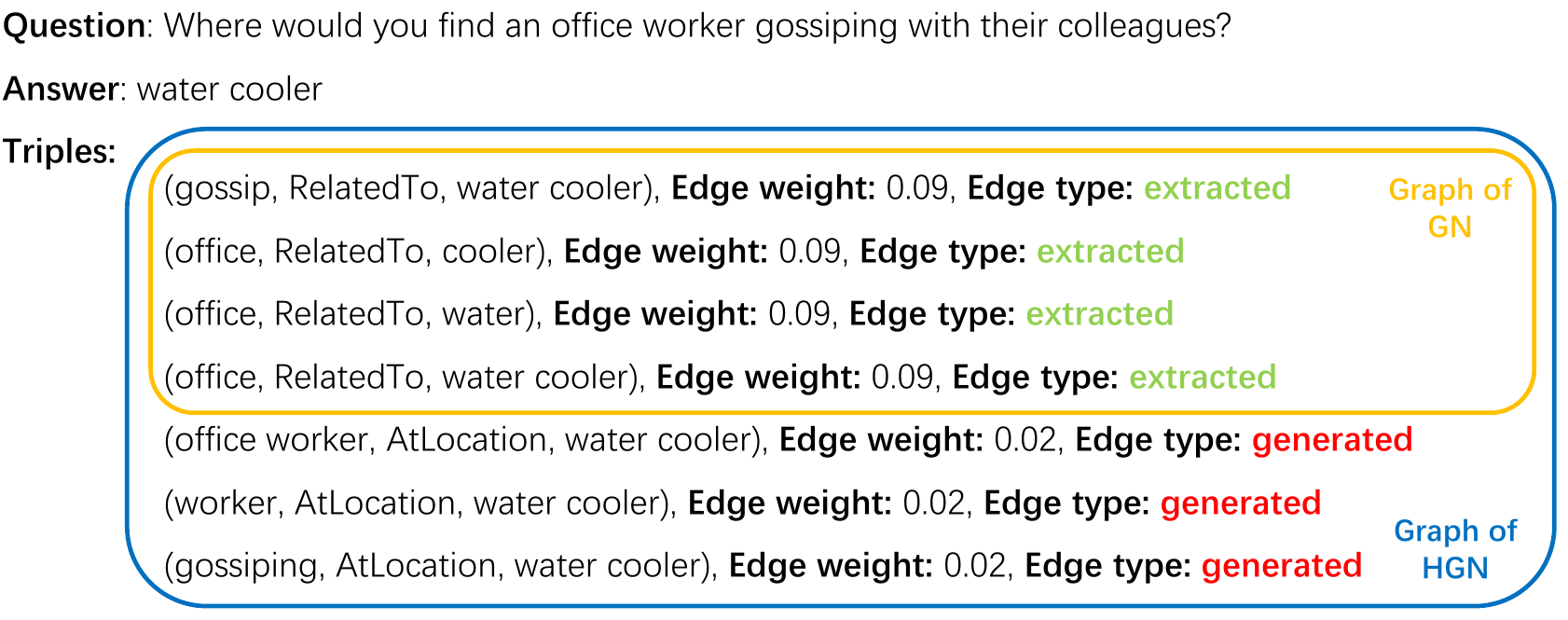}%
}
\caption{\textbf{Representative cases from the development set of CommonsenseQA.}}
\label{fig:cases}
\end{figure*}

In addition to the experiments in \sectionref{sec:experiments}, we present a case study here, which compares a HGN-generated graph with a KG-extracted graph used by GN.
On the development set of CommonsenseQA, there are two dominating cases and we show the representative instance of each one.
Figure~\ref{fig:cases} (a) shows the first case, where HGN prunes edges from the extracted graph. Our HGN assigns the highest weights to the most helpful facts (\texttt{book}, \texttt{AtLocation}, \texttt{house}), (\texttt{telephone book}, \texttt{AtLocation}, \texttt{house}). It also downweight unhelpful fact (\texttt{place}, \texttt{IsA}, \texttt{house}) and invalid fact (\texttt{usually}, \texttt{RelatedTo}, \texttt{house}).
Figure~\ref{fig:cases} (b) shows the second case, where new generated facts are incorporated into reasoning. All generated facts that are kept by the model make sense in the context and help identify the answer. Both cases suggest that our model improve the quality of the contextualized knowledge graph compared to the current methods that only rely on extracted facts.

